\documentclass[10pt,twocolumn,letterpaper]{article}

\usepackage{cvpr}
\usepackage{times}
\usepackage{epsfig}
\usepackage{graphicx}
\usepackage{amsmath}
\usepackage{amssymb}
\usepackage{subfigure}
\usepackage{indentfirst}
\usepackage{bm}
\usepackage{multirow}

\cvprfinalcopy 

\def\cvprPaperID{2579} 

\ifcvprfinal\fi
\begin{document}

\title{Adaptive Graph Convolutional Network with Attention Graph Clustering for Co-saliency Detection}
\author{Kaihua Zhang$^1$, Tengpeng Li$^1$, Shiwen Shen$^2$, Bo Liu$^2$\thanks{Corresponding author. This work is supported in part by National Major Project of China for New Generation of AI (No. 2018AAA0100400), in part by the NSFC (61876088, 61825601), in part by the NSF of Jiangsu Province (BK20170040).}, Jin Chen$^1$,Qingshan Liu$^1$\\
$^1$B-DAT and CICAEET, Nanjing University of Information
Science and Technology, Nanjing, China\\
$^2$JD Digits, Mountain View, CA, USA\\
{\tt\small \{zhkhua,kfliubo\}@gmail.com}
}

\maketitle

\begin{abstract}
Co-saliency detection aims to discover the common and salient foregrounds from a group of relevant images. For this task, we present a novel adaptive graph convolutional network with attention graph clustering (GCAGC). Three major contributions have been made, and are experimentally shown to have substantial practical merits. First, we propose a graph convolutional network design to extract information cues to characterize the intra- and inter-image correspondence. Second, we develop an attention graph clustering algorithm to discriminate the common objects from all the salient foreground objects in an unsupervised fashion. Third, we present a unified framework with encoder-decoder structure to jointly train and optimize the graph convolutional network, attention graph cluster, and co-saliency detection decoder in an end-to-end manner. We evaluate our proposed GCAGC method on three co-saliency detection benchmark datasets (iCoseg, Cosal2015 and COCO-SEG). Our GCAGC method obtains significant improvements over the state-of-the-arts on most of them.
\end{abstract}
\section{Introduction}
Human is able to exhibit visual fixation to attend to the attractive and interesting regions and objects for future processing \cite{cong2018review}. Co-saliency detection model simulates the human visual system to perceive the scene, and searches for the common and salient foregrounds in an image group. Co-saliency has been used in various applications to improve the understanding of image/video content, such as image/video co-segmentation~\cite{tsai2018image, fu2015object, fu2014object, wang2017video}, imge/video salient object detection~\cite{fan2019shifting,Zhang2020UCNet,Fu2020JLDCF,zhao2019contrast}, object co-localization~\cite{tang2014co}, and image retrieval~\cite{yang2011object}.
%
%
%

In co-saliency detection, the semantic categories of the common salient objects are unknown. Thus, the designed algorithm needs to infer such information from the specific content of the input image group. Therefore, the co-saliency detection algorithm design usually focuses on addressing two key challenges: (1) extracting informative image feature representations to robustly describe the image foregrounds; and (2) designing effective computational frameworks to formulate and detect the co-saliency. Conventional hand-engineered features, such as Gabor filters, color histograms and SIFT descriptors \cite{lowe2004distinctive} have been widely used in many co-saliency detection methods~\cite{fu2013cluster,liu2013co, ye2015co}. However, hand-crafted shallow features usually lack the ability to fully capture the large variations of common object appearances, and complicated background textures \cite{wang2019robust}.  Recently, researchers improve co-saliency detection using deep-learning-based high-level feature representations, and have shown promising results ~\cite{zhang2016co, zhang2015detection, zhang2019co}. Nonetheless, these approaches separate the representation extraction from co-saliency detection as two distinct steps, and lose the ability to tailor the image features towards inferring co-salient regions \cite{hsu2018unsupervised}. End-to-end algorithms adopting convolutional neural networks (CNNs) \cite{hsu2018unsupervised, wang2019robust, wei2017group} have been developed to overcome this problem, and demonstrated state-of-the-art performance. Although CNN is able to extract image representations in a data-driven way, it is the sub-optimal solution to model long-range dependencies \cite{wang2018non}. CNN captures long-range dependencies by deeply stacking convolutional operations to enlarge the receptive fields. However, the repeated convolutional operations cause optimization difficulties \cite{wang2018non, he2016deep}, and make multi-hop dependency modeling \cite{wang2018non}. Moreover, it becomes even more challenging for the CNN to accurately modeling the inter-image non-local dependencies for the co-salient regions in the image group.

To address the aforementioned challenges, we develop a novel adaptive graph convolutional network with attention graph clustering (GCAGC) for co-saliency detection. We first utilize a CNN encoder to extract multi-scale feature representations from the image group, and generate combined dense feature node graphs. We then process the dense graphs with the proposed adaptive graph convolutional network (AGCN). Compared with only depending on the progressive behavior of the CNN, the AGCN is able to capture the non-local and long-range correspondence directly by computing the interactions between any two positions of the image group, regardless of their intra- and inter-image positional distance. The output from AGCN is further refined by an attention graph clustering module (AGCM) through generated co-attention maps. A CNN decoder is employed in the end to output the finally predicted co-saliency maps. A unified framework is designed to jointly optimize all the components together.

The main contributions of this paper are threefold:
\begin{itemize}

\item We provide an adaptive graph convolutional network design to simultaneously capture the intra- and inter-image correspondence of an image group. Compared with conventional approaches, this AGCN directly computes the long-range interactions between any two image positions, thus providing more accurate measurements.

\item We develop an attention graph clustering module to differentiate the common objects from salient foregrounds. This AGCM is trained in an unsupervised fashion, and generates co-attention maps to further refine the estimated co-salient foregrounds.

\item We present an end-to-end computational framework with encoder-decoder CNN structure to jointly optimize the graph clustering task and the co-saliency detection objective, while learning adaptive graph dependencies.
\end{itemize}

\section{Related Work}
\textbf{Image Co-saliency Detection.} This task identifies common distinct foregrounds and segments them from multiple images. Various strategies have been developed for this task. Bottom-up approaches first score each pixel/sub-region in the image group, and then combine similar regions in a bottom-up fashion. Hand-crafted features~\cite{fu2013cluster,ge2016co,li2015efficient,liu2013co,ye2015co} or deep-learning-based features \cite{zhang2016co, zhang2016detection} are usually employed to score such sub-regions. Fu \textit{et al.}~\cite{fu2013cluster} utilize three visual attention priors in a cluster-based framework. Liu \textit{et al.}~\cite{liu2013co} define background and foreground cues to capture the intra- and inter-image similarities. Pre-trained CNN and restricted Boltzmann machine are used in \cite{zhang2016co} and \cite{zhang2016detection} to extract information cues to detect common salient objects, respectively. In contrast, fusion-based algorithms \cite{tsai2019deep, cao2014self, jerripothula2016image} are designed to discover useful information from the predicted results generated by several existing saliency or co-saliency detection methods. These methods fuse the detected region proposals by region-wise adaptive fusion~\cite{jerripothula2016image}, adaptive weight fusion~\cite{cao2014self} or stacked-autoencoder-enabled fusion~\cite{tsai2019deep}. Learning-based methods are the third category of co-saliency detection algorithms, and developed to learn the co-salient pattern directly from the image group. In~\cite{hsu2018unsupervised}, an unsupervised CNN with two graph-based losses is proposed to learn the intra-image saliency and cross-image concurrency, respectively. Zhang \textit{et al.}~\cite{zhang2019co} design a hierarchical framework to capture co-salient area in a mask-guided fully CNN. Wei \textit{et al.}~\cite{wei2017group} design a multi-branch architecture to discover the interaction across images and the salient region in single image simultaneously. A semantic guided feature aggregation architecture is proposed to capture the concurrent and fine-grained information in~\cite{wang2019robust}. Although many methods have been developed, this field still lacks of research on addressing the limitations of CNN for capturing long-range intra- and inter-image dependencies.

\textbf{Graph Neural Networks (GNNs).} GNNs~\cite{gori2005new, scarselli2008graph} are the models for capturing graph dependencies via message passing between the nodes of graphs. Different from standard neural network, GNNs retain a state that can represent information from its neighborhood with arbitrary depth~\cite{zhou2018graph}. Convolutional graph neural networks (GCNs)~\cite{bruna2013spectral,defferrard2016convolutional,kipf2016semi,levie2018cayleynets, atwood2016diffusion,niepert2016learning,gilmer2017neural} are a variant of GNNs, and aim to generalize convolution to graph domain. Algorithms in this direction are often categorized as the spectral-based approaches~\cite{bruna2013spectral,defferrard2016convolutional,kipf2016semi,levie2018cayleynets}, and the spatial-based approaches~\cite{atwood2016diffusion,niepert2016learning,gilmer2017neural}. The former ones work with a spectral representation of the graphs; and the latter ones define the operation directly on graph, and extract information from groups of spatially connected neighbours. Recently, GNN and GCN have demonstrated promising results in various computer vision tasks, including scene graph generation~\cite{yang2018graph, li2018factorizable, gu2019scene}, point clouds classification and segmentation~\cite{landrieu2018large, wang2019dynamic}, semantic segmentation~\cite{wang2019zero, qi20173d}, action recognition~\cite{yan2018spatial} and visual reasoning and question answering~\cite{chen2018iterative, narasimhan2018out}. More comprehensive review of GNNs can be found in \cite{zhou2018graph, wu2019comprehensive}.

\begin{figure*}[t]
\begin{center}
\begin{tabular}{c}
\includegraphics[width=1\linewidth]{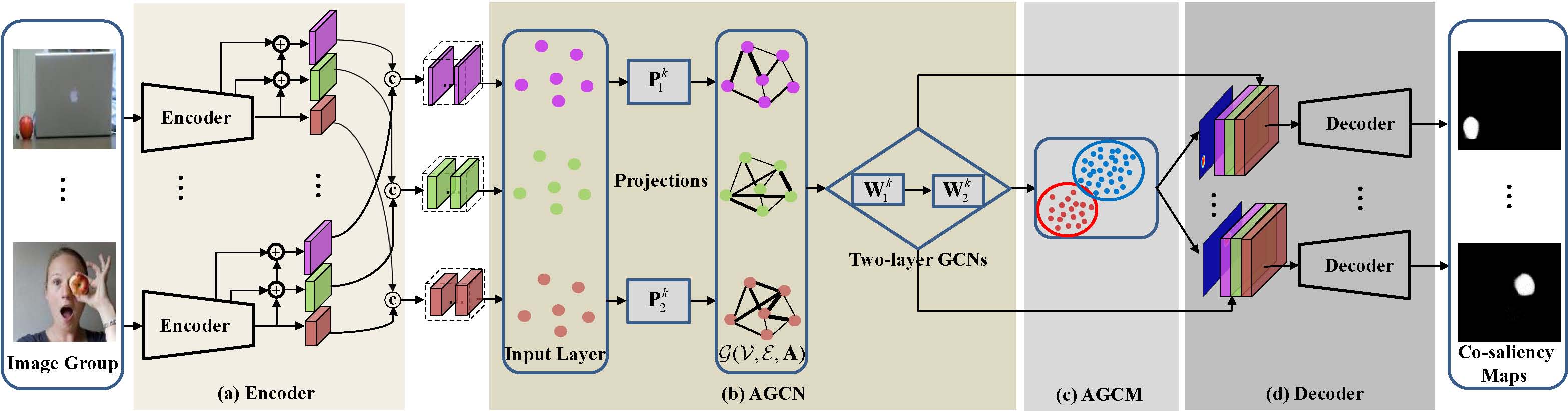}
\end{tabular}
\end{center}
   \caption{Pipeline of the proposed GCAGC for co-saliency detection. Given a group of images as input, we first leverage a backbone CNN as encoder (a) to extract the multi-scale features of each image, and then we adopt the feature pyramid network (FPN)~\cite{lin2017feature} to fuse all the image features from top to down. Next, the lateral output features as node representations are fed into the AGCN (b). The output features of AGCN via two-layer GCNs are then fed into the AGCM (c), generating a set of object co-attention maps.
   Finally, the co-attention maps and the output features of AGCN are concatenated and fed into the decoder (d), producing corresponding co-saliency maps.
   \textcircled{+}: element-wise addition; \textcircled{c}: concatenation; $\mathcal{G}(\mathcal{V},\mathcal{E},\textbf{A})$: graph of nodes $\mathcal{V}$, edges $\mathcal{E}$ and adjacency matrix $\textbf{A}$; $\textbf{P}^k_1$, $\textbf{P}^k_2$: learnable projection matrices for graph learning; $\textbf{W}_1^k$ and $\textbf{W}_2^k$: learnable weight matrices in the adopted two-layer GCNs.}
\label{fig:pipeline}
\end{figure*}
\textbf{Graph Clustering.} This task divides the graph nodes into related groups. Early works~\cite{girvan2002community,sun2009itopicmodel} develop shallow approaches for graph clustering. Girvan~\textit{et al.}~\cite{girvan2002community} use centrality indices to discover boundaries of different nodes groups. Wang~\textit{et al.}~\cite{wang2017community}  develop a modularized non-negative matrix factorization approach to incorporate the community structure into the graph embedding, and then perform traditional clustering methods on the embedded features. The limitations of these works are that they only handle partial graph structure or shallow relationships between the content and the structure data~\cite{wang2019attributed}. In contrast, deep-learning-based approaches~\cite{pan2019learning,wang2019attributed} are developed recently to improve graph clustering. Pan \textit{et al.}~\cite{pan2019learning} present an adversarially regularized framework to extract the graph representation to perform graph clustering. Wang \textit{et al.}~\cite{wang2019attributed} develop a goal-directed deep learning approach to jointly learn graph embedding and graph clustering together. More detailed review of graph clustering is provided in \cite{bader2013graph}.
\section{Proposed Approach}
%
\subsection{Method Overview}\label{sec:problem formulation}
Given a group of $N$ relevant images $\mathcal{I}=\{\textbf{\textit{I}}^n\}_{n=1}^N$, the task of co-saliency detection aims to highlight the shared salient foregrounds against backgrounds, predicting the corresponding response maps $\mathcal{M}=\{\textbf{M}^n\}_{n=1}^N$.
To achieve this goal, we learn a deep GCAGC model to predict $\mathcal{M}$ in an end-to-end fashion.
%

%
Figure~\ref{fig:pipeline} illustrates the pipeline of our approach, which consists of four key components: (a) Encoder, (b) AGCN, (c) AGCM and (d) Decoder.
Specifically, given input $\mathcal{I}$, we first adopt the VGG16 backbone network~\cite{simonyan2014very} as the encoder to extract their features by removing the fully-connected layers and softmax layer.
Afterwards, we leverage the FPN~\cite{lin2017feature} to fuse the features of pool3, pool4 and pool5 layers, generating three lateral intermediate feature maps $\mathcal{X}=\{{\textbf{X}}^k\}_{k=1}^3$ as the multi-scale feature representations of $\mathcal{I}$,
%
%
Then, for each ${\textbf{X}}^k\in \mathcal{X}$, we design a sub-graph $\mathcal{G}^k$ with a learnable structure that is adaptive to our co-saliency detection task, which is able to well capture the long-range intra- and inter-image correspondence while preserving the spatial consistency of the saliency.
Meanwhile, to fully capture multi-scale information for feature enhancement, the sub-graphs are combined into a multi-graph $\mathcal{G}=\cup_{k}\mathcal{G}^k$.
Then, $\mathcal{G}$ is integrated into a simple two-layer GCNs $\mathcal{F}_{gcn}$~\cite{kipf2016semi}, generating the projected GC filtered features $\mathcal{F}_{gcn}(\mathcal{X})=\{\mathcal{F}_{gcn}(\textbf{X}^k)\}_{k=1}^3$.
Recent works~\cite{li2018deeper,li2019label} show that the GC filtering of GCNs~\cite{kipf2016semi} is a Laplacian smoothing process, and hence it makes the salient foreground features of the same category similar, thereby well preserving spatial consistency of the foreground saliency, which facilitates the subsequent intra- and inter-image correspondence.
Afterwards, $\mathcal{F}_{gcn}(\mathcal{X})$ are fed into a graph clustering module $\mathcal{F}_{gcm}$, producing a group of co-attention maps $\textit{\textbf{M}}_{catt}$, which help to further refine the predicted co-salient foregrounds while suppressing the noisy backgrounds.
Finally, the concatenated features $\textit{\textbf{M}}_{catt}\copyright\mathcal{F}_{gcn}(\mathcal{X})$ are fed into a decoder layer, producing the finally predicted co-saliency maps.

%
%

\subsection{Adaptive Graph Convolution Network}\label{sec:AGCN}
As aforementioned, the AGCN is to process features as Laplacian smoothing~\cite{li2018deeper} that can benefit long-range intra- and inter-image correspondence while preserving spatial consistency.
Numerous graph based works for co-saliency detection~\cite{hsu2018unsupervised,tsai2018image,zheng2018feature,hsu2019deepco3,jiang2019unified,li2019co} have been developed to better preserve spatial consistency, but they perform intra-saliency detection and inter-image correspondence independently, which cannot well capture the interactions between co-salient regions across images that are essential to co-saliency detection, thereby leading to sub-optimal performance.
Differently, our AGCN constructs a dense graph that takes all input image features as the node representations. Meanwhile, each edge of the graph models the interactions between any pair-wise nodes regardless of their positional distance, thereby well capturing long-range dependencies.
Hence, both intra-saliency detection and inter-image correspondence can be jointly implemented via feature propagation on the graph under a unified framework without any poster-processing, leading to a more accurate co-saliency estimation than those individually processing each part~\cite{hsu2018unsupervised,tsai2018image,zheng2018feature,hsu2019deepco3,jiang2019unified,li2019co}.

%
%
\begin{figure}[t]
\begin{center}
\begin{tabular}{l}
\includegraphics[width=0.85\linewidth]{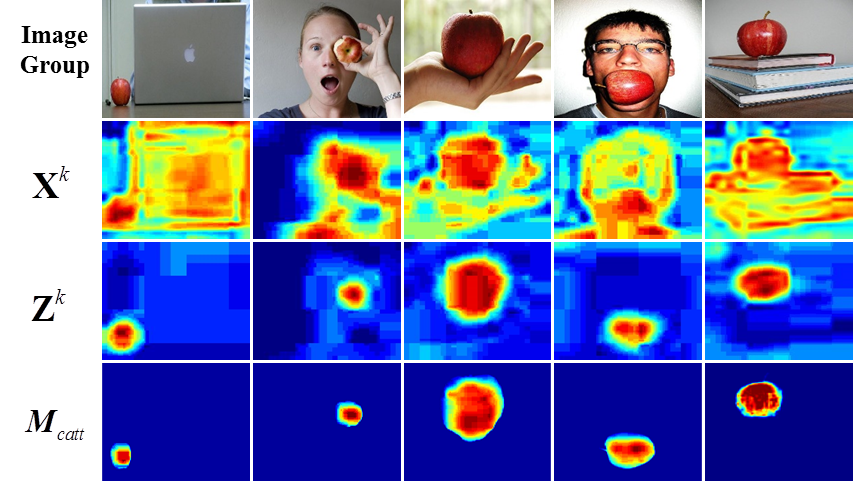}
\end{tabular}
\end{center}
\caption{Illustration of the effect of GC filtering. The GC filtered signal projections $\textbf{Z}^k$ preserve better spatial consistency of the salient foregrounds than the input graph signals $\textbf{X}^k$ that highlight more noisy backgrounds. Afterwards, the co-attention maps $\textit{\textbf{M}}_{catt}$ generated by our AGCM in~\S~\ref{sec:AGCM} further reduce the noisy backgrounds existing in $\textbf{Z}^k$.}
\label{fig:heatmap}
\end{figure}

\textbf{Notations of Graph.}
We construct a multi-graph $\mathcal{G}(\mathcal{V},\mathcal{E},\textbf{A})=\cup_{k=1}^3\mathcal{G}^k(\mathcal{V}^k,\mathcal{E}^k,\textbf{A}^k)$ that is composed of three sub-graphs $\mathcal{G}^k$, where node set $\mathcal{V}=\{\mathcal{V}^k\}$, edge set $\mathcal{E}=\{\mathcal{E}^k\}$, adjacent matrix $\textbf{A}=\sum_k\textbf{A}^k$, $\mathcal{V}^k=\{v^k_i\}$ denotes the node set of $\mathcal{G}^k$ with node $v_i^k$, $\mathcal{E}^k=\{e^k_{ij}\}$ denotes its edge set with edge $e^k_{ij}$, $\mathbf{A}^k$ denotes its adjacent matrix, whose entry $\mathbf{A}^k(i,j)$ denotes the weight of edge $e^k_{ij}$.
${\textbf{X}}^k=[\textit{\textbf{x}}_1^k,\ldots,\textit{\textbf{x}}_{Nwh}^k]^\top$ denotes the feature matrix of $\mathcal{G}^k$, where $\textit{\textbf{x}}_i^k\in \mathbb{R}^{d^k}$ is the features of node $v_i^k$ with dimension $d^k$.
%

\textbf{Adjacency Matrix $\textbf{A}$.}
The vanilla GCNs~\cite{kipf2016semi} construct a fixed graph without training, which cannot guarantee to be best suitable to a specific task~\cite{henaff2015deep}.
Recently, some works~\cite{henaff2015deep,li2018adaptive,jiang2019unified} have investigated adaptive graph learning techniques through learning a parameterized adjacency matrix tailored to a specific task.
Inspired by this and the self-attention mechanism in~\cite{wang2018non}, for sub-graph $k$, to learn a task-specific graph structure, we define a learnable adjacency matrix as
\begin{equation}
\textbf{A}^k=\sigma({\textbf{X}}^k\textbf{P}^k_1({\textbf{X}}^k\textbf{P}^k_2)^\top),
\label{eq:Ak}
\end{equation}
where $\sigma(x)=\frac{1}{1+e^{-x}}$ denotes the sigmoid function, $\textbf{P}^k_1, \textbf{P}^k_2\in \mathbb{R}^{d^k\times r}$ are two learnable projection matrices that reduce the dimension of the node features from $d^k$ to $r<d^k$.

To combine multiple graphs in GCNs, as in~\cite{wang2018videos}, we simply element-wisely add the adjacency matrices of all $\mathcal{G}^k$ to construct the adjacency matrix of $\mathcal{G}$ as
\begin{equation}
\label{eq:A}
\textbf{A}=\textbf{A}^1+\textbf{A}^2+\textbf{A}^3.
\end{equation}
\textbf{Graph Convolutional Filtering.}
We employ the two-layer GCNs proposed by~\cite{kipf2016semi} to perform graph convolutions as
\begin{equation}
\begin{aligned}
\textbf{Z}^k&= \mathcal{F}_{gcn}(\textbf{X}^k)\\
&= \mathcal{F}_{{softmax}}(\hat{\textbf{A}}\mathrm{ReLU} (\mathcal{F}_{gcf}({\hat{\textbf{A}}},\textbf{X}^k)\textbf{W}_1^k) \textbf{W}_2^k),
\end{aligned}
\label{eq:GCF}
\end{equation}
where the GC filtering function is defined as~\cite{li2019label}
\begin{equation}
\label{eq:gcf}
\mathcal{F}_{gcf}({\hat{\textbf{A}}},\textbf{X}^k)=\hat{\textbf{A}}\textbf{X}^k,
\end{equation}
$\textbf{W}_1^k\in \mathbb{R}^{d^k\times c_1^k}$, $\textbf{W}_2^k\in \mathbb{R}^{c_1^k\times c^k}$ denote the learnable weight matrices of two fully-connected layers for feature projections, $\hat{\textbf{A}}=\tilde{\textbf{D}}^{-\frac{1}{2}}\tilde{\textbf{A}}\tilde{\textbf{D}}^{-\frac{1}{2}}$,
where $\tilde{\textbf{A}}=\textbf{A}+\textbf{I}$, $\textbf{A}$ is defined by (\ref{eq:A}) and $\textbf{I}$ denotes the identity matrix, $\tilde{\textbf{D}}(i,i)=\sum_j\tilde{\textbf{A}}(i,j)$ is the degree matrix of $\tilde{\textbf{A}}$ that is diagonal.
Recent work~\cite{li2019label} has shown that the GC filtering $\mathcal{F}_{gcf}$ (\ref{eq:gcf}) is low-pass and hence it can make the output signal projections $\textbf{Z}^k$ smoother in the same cluster, so as to well preserve the spatial consistency of the salient foregrounds across images as illustrated by Figure~\ref{fig:heatmap}.
%
However, some intra-consistency but non-salient regions have also been highlighted.
To overcome this issue, in the following section, we will
present an attention graph clustering technique to further refine $\textbf{Z}^k$ to focus on co-salient regions.
\begin{figure}
\begin{center}
\begin{tabular}{l}
\includegraphics[width=1\linewidth]{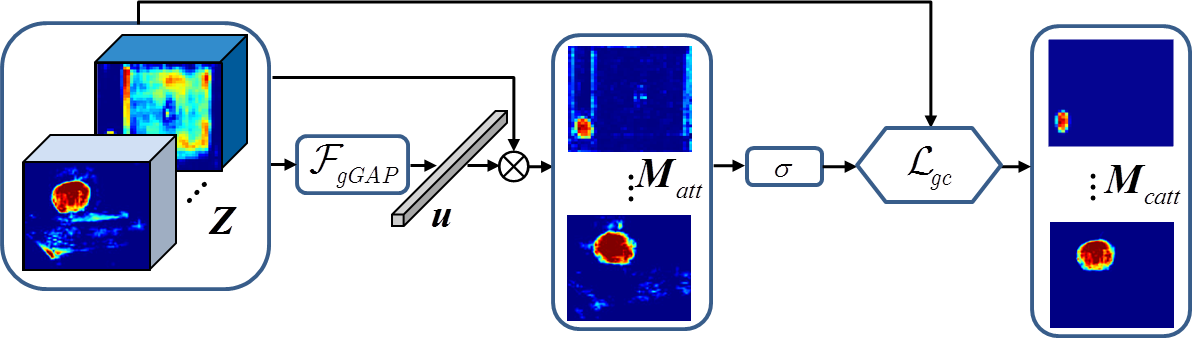}
\end{tabular}
\end{center}
\caption{The schematic diagram of our AGCM $\mathcal{F}_{gcm}$. Please refer to the text part for details.}
\label{fig:AGCM}
\end{figure}
\subsection{Attention Graph Clustering Module}
\label{sec:AGCM}
Figure~\ref{fig:AGCM} shows the schematic diagram of our AGCM $\mathcal{F}_{gcm}$.
Specifically, given the GC filtering projections $\textbf{Z}^k\in\mathbb{R}^{Nwh\times c^k}, k=1,2,3$ in (\ref{eq:GCF}), we obtain a multi-scale feature matrix by concatenating them as $\textbf{Z}=[\textbf{Z}^1,\textbf{Z}^2,\textbf{Z}^3]=[\textit{\textbf{z}}_1,\ldots,\textit{\textbf{z}}_{Nwh}]^\top\in \mathbb{R}^{Nwh\times d}$, where the multi-scale node features $\textit{\textbf{z}}_i\in \mathbb{R}^d$, $d=\sum_kc^k$. Next, we reshape $\textbf{Z}$ to tensor $\textit{\textbf{Z}}\in\mathbb{R}^{N\times w\times h\times d}$ as input of $\mathcal{F}_{gcm}$. Then, we define a group global average pooling (gGAP) function $\mathcal{F}_{gGAP}$ as
\begin{equation}
\label{eq:gGAP}
\textit{\textbf{u}}=\mathcal{F}_{gGAP}(\textit{\textbf{Z}})=\frac{1}{Nwh}\sum_{n,i,j}\textit{\textbf{Z}}(n,i,j,:),
\end{equation}
which  outputs a global statistic feature $\textit{\textbf{u}} \in \mathbb{R}^d$ as the multi-scale semantic saliency representation that encodes the global useful group-wise context information.
Afterwards, we correlate $\textit{\textbf{u}}$ and $\textit{\textbf{Z}}$ to generate a group of attention maps that can fully highlight the intra-saliency:
\begin{equation}
\label{eq:coattention}
\textit{\textbf{M}}_{att}=\textit{\textbf{u}}\otimes\textit{\textbf{Z}},
\end{equation}
where $\textit{\textbf{M}}_{att} \in \mathbb{R}^{N\times w\times h}$, $\otimes$ denotes correlation operator.
Then, we use sigmoid function $\sigma$ to re-scale the values of $\textit{\textbf{M}}_{att}$ to $[0,1]$ as
\begin{equation}
\label{eq:weightW}
\textit{\textbf{W}}=\sigma(\textit{\textbf{M}}_{att}).
\end{equation}
From Figure~\ref{fig:AGCM}, we can observe that $\textit{\textbf{M}}_{att}$ discovers intra-saliency that preserves spatial consistency, but some noisy non-co-salient foregrounds have also been highlighted.
To alleviate this issue, we exploit an attention graph clustering technique to further refine the attention maps, which are able to better differentiate the common objects from salient foregrounds.
Motivated by the weighted kernel $k$-means approach in~\cite{dhillon2007weighted}, we define the objective function of AGCM as
\begin{equation}
\begin{split}
\mathcal{L}_{gc}= \sum_{\textit{\textbf{z}}_i \in \pi_{f}} w_{i} \|\textit{\textbf{z}}_i-\textit{\textbf{m}}_{f}\|^2
+\sum_{\textit{\textbf{z}}_i \in \pi_{b}} w_{i} \|\textit{\textbf{z}}_i-\textit{\textbf{m}}_{b}\|^2,\\
\end{split}
\label{eq:weighted clustering}
\end{equation}
where $\pi_f$ and $\pi_b$ denote the clusters of foreground and background respectively, $\textit{\textbf{m}}_{f}=\frac{\sum_{\textit{\textbf{z}}_i\in\pi_f}\textit{\textbf{z}}_iw_i}{\sum_{\textit{\textbf{z}}_i\in\pi_f}w_i}$ and similar for $\textit{\textbf{m}}_{b}$, $w_i$ denotes the $i$-th element of $\textit{\textbf{W}}$ in (\ref{eq:weightW}).

Following~\cite{dhillon2007weighted}, we can readily show that the minimization of the objective $\mathcal{L}_{gc}$ in (\ref{eq:weighted clustering}) is equivalent to
\begin{equation}
\label{eq:lgc}
\min_{{{\textbf{Y}}}}\{\mathcal{L}_{gc}=-\mathrm{trace}({{{\textbf{Y}}}}^\top{\textbf{K}}{{\textbf{Y}}})\},
\end{equation}
where ${\textbf{K}}=\textbf{D}^\frac{1}{2}\textbf{ZZ}^\top \textbf{D}^\frac{1}{2}$, $\textbf{D}=\mathrm{diag}({w}_1,\ldots,{w}_{Nwh})$, ${{\textbf{Y}}}\in \mathbb{R}^{Nwh\times 2}$
satisfies $\textbf{Y}^\top \textbf{Y}=\textbf{I}$.
%

Let $\textit{\textbf{y}}\in \{0,1\}^{Nwh}$ denote the indictor vector of the clusters, and $\textit{\textbf{y}}(i)=1$ if $i\in\pi_f$, else, $\textit{\textbf{y}}(i)=0$.
We choose ${\textbf{Y}}=[\textit{\textbf{y}}/{\sqrt{|\pi_f|}},(\mathbf{1}-\textit{\textbf{y}})/{\sqrt{|\pi_b|}}]$ that satisfies $\textbf{Y}^\top \textbf{Y}=\textbf{I}$ and put it into (\ref{eq:lgc}), yielding the loss function of our AGCM
\begin{equation}
\label{eq:lossAGCM}
\mathcal{L}_{gc}=-\left(\frac{\textit{\textbf{y}}^\top\textbf{K}\textit{\textbf{y}}}{\textit{\textbf{y}}^\top\textit{\textbf{y}}}+\frac{(\textbf{1}-\textit{\textbf{y}})^\top\textbf{K}(\textbf{1}-\textit{\textbf{y}})}{(\textbf{1}-\textit{\textbf{y}})^\top(\textbf{1}-\textit{\textbf{y}})}\right).
\end{equation}
Now, we show the relationship between the above loss $\mathcal{L}_{gc}$ and graph clustering. We first construct the graph of GC as $\mathcal{G}_{gc}(\mathcal{V}_{gc},\mathcal{E}_{gc},\textbf{K})$, which is made up of node set $\mathcal{V}_{gc}=\mathcal{V}_f\cup \mathcal{V}_b$,
where $\mathcal{V}_f$ is the set of foreground nodes and $\mathcal{V}_b$ is the set of background nodes, $\mathcal{E}_{gc}$ denotes the edge set such that the weight of edge between nodes $i$ and $j$ is equal to $\textbf{K}(i,j)$, where $\textbf{K}$ is its adjacency matrix defined in (\ref{eq:lgc}).
Let us denote $\mathrm{links}(\mathcal{V}_l,\mathcal{V}_l)=\sum_{i\in\mathcal{V}_l,j\in \mathcal{V}_l}\textbf{K}(i,j), l=f,b$,
then, it is easy to show that minimizing $\mathcal{L}_{gc}$  (\ref{eq:lossAGCM}) is equivalent to maximizing the ratio association objective~\cite{shi2000normalized} for graph clustering task
\begin{equation}
\max\left\{\sum_{l=f,g}\frac{\mathrm{links}(\mathcal{V}_l,\mathcal{V}_l)}{|\mathcal{V}_l|}\right\}.
\end{equation}
where $|\mathcal{V}_l|$ denotes the cardinality of set $\mathcal{V}_l$.

Directly optimizing $\mathcal{L}_{gc}$ (\ref{eq:lossAGCM}) yields its continuous relaxed solution $\hat{\textit{\textbf{y}}}$. Then, we reshape $\hat{\textit{\textbf{y}}}$ into a group of $N$ co-attention maps ${\textit{\textbf{M}}}_{catt}\in\mathbb{R}^{N\times w\times h}$.
Finally, the learned co-attention maps ${\textit{\textbf{M}}}_{catt}$ and the input features $\textit{\textbf{Z}}\in\mathbb{R}^{N\times w\times h\times d}$ of the AGCM are concatenated, yielding the enhanced features $\textit{\textbf{F}}\in \mathbb{R}^{N\times w\times h\times (d+1)}$:
\begin{equation}
\label{eq:F}
\textit{\textbf{F}}={\textit{\textbf{M}}}_{catt}\copyright\textit{\textbf{Z}},
\end{equation}
where $\copyright$ denotes concatenation operator, which serves as the input of the following decoder network.
\begin{figure*}
\begin{center}
\begin{tabular}{c}
\includegraphics[width=1 \textwidth]{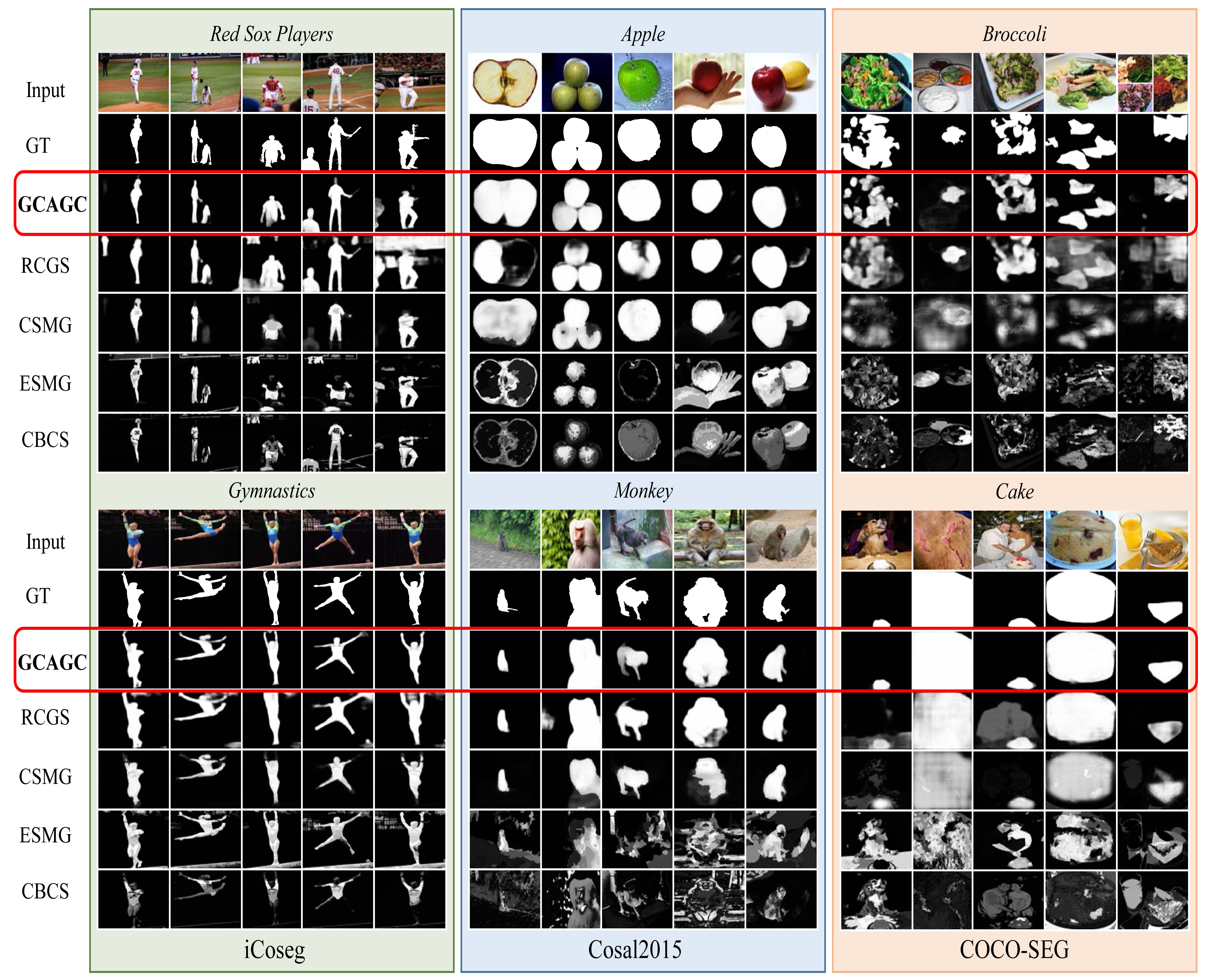}
\end{tabular}
\end{center}
   \caption{Visual comparisons of our GCAGC method compared with other state-of-the-arts, including CBCS~\cite{fu2013cluster}, ESMG~\cite{li2015efficient}, CSMG~\cite{zhang2019co} and RCGS~\cite{wang2019robust}.}
\label{fig:Comparison1}
\end{figure*}
\subsection{Decoder Network} \label{sec:decoder}
Our decoder network has an up-sampling module that is consist of a $3\times 3$ convolutional layer to decrease feature channels, a ReLU layer and a deconvolutional layer with stride $=2$ to enlarge resolution.
Then, we repeat this module three times until reaching the finest resolution for accurate co-saliency map estimation, following a $1\times 1$ convolutional layer and a sigmoid layer to produce a group of co-saliency map estimations. 
Given the features $\textit{\textbf{F}}$ computed by (\ref{eq:F}) as input, the decoder network generates a group of co-saliency maps $\mathcal{M}=\{\textbf{M}^n\in \mathbb{R}^{w\times h}\}_{n=1}^N$.
We then leverage a weighted cross-entropy loss for pixel-wise classification
\begin{equation}
\begin{split}
\mathcal{L}_{cls} = - \frac{1}{P\times N} \sum_{n=1}^{N} \sum_{i=1}^{P} \{ \rho^n \textbf{M}^n(i)\log(\textbf{M}_{gt}^n(i))\\
 - (1-\rho^n) (1-\textbf{M}^n(i))\log(1-\textbf{M}_{gt}^n(i)) \},
\label{eq:clsloss}
\end{split}
\end{equation}
where $\textbf{M}_{gt}^n$ denotes the ground-truth mask of image $\textit{\textbf{I}}^n\in\mathcal{I}$, $P$ denotes the pixel number of image $\textit{\textbf{I}}^n$ and $\rho^{n}$ denotes the ratio of all positive pixels over all pixels in image $\textit{\textbf{I}}^{n}$.

All the network parameters are jointly learned by minimizing the following multi-task loss function
\begin{equation}
\label{eq:loss}
\mathcal{L}= \mathcal{L}_{cls} + \lambda\mathcal{L}_{gc},
\end{equation}
where $\mathcal{L}_{gc}$ is the attention graph clustering loss defined by (\ref{eq:lossAGCM}), $\lambda>0$ is a trade-off parameter.
We train our network by minimizing $\mathcal{L}$ in an end-to-end manner, and the learned GCAGC model is directly applied to processing input image group, predicting the corresponding co-saliency maps without any post-processing.

\section{Results and Analysis}
\subsection{Implementation Details}
\label{sec:implementation}
The training of our GCAGC model includes two stages:

\textbf{Stage 1.} For fair comparison, we adopt the VGG16 network~\cite{simonyan2014very} as the backbone network, which is pre-trained on the ImageNet classification task~\cite{deng2009imagenet}.
Following the input settings in~\cite{wei2017group,wang2019robust}, we randomly select $N=5$ images as one group from one category and then select a mini-batch groups from all categories in the COCO dataset~\cite{lin2014microsoft}, which are sent into the network at the same time during training.
All the images are resized to the same size of $224 \times 224$ for easy processing.
The model is optimized by the Adam algorithm~\cite{kingma2014adam} with a weight decay of 5e-4 and an initial learning rate of 1e-4 which is reduced by a half every $25,000$ iterations. This training process converges until $100,000$ iterations.

\textbf{Stage 2.} We further fine-tune our model using MSRA-B dataset~\cite{liu2011learning} to better focus on the salient areas.
All the parameter settings are the same as those in  \textbf{Stage 1} except for the training iterations =$10,000$.
Note that when training, to match the size of input group, we augment the single salient image to $N=5$ different images as a group using affine transformation, horizontal flipping and left-right flipping.

During testing, we divide all images into several mini-groups to produce the final co-saliency map estimations.
The network is implemented in PyTorch with a RTX 2080Ti GPU for acceleration.
\begin{figure*}
\begin{center}
\begin{tabular}{c}
\includegraphics[width=.98\linewidth]{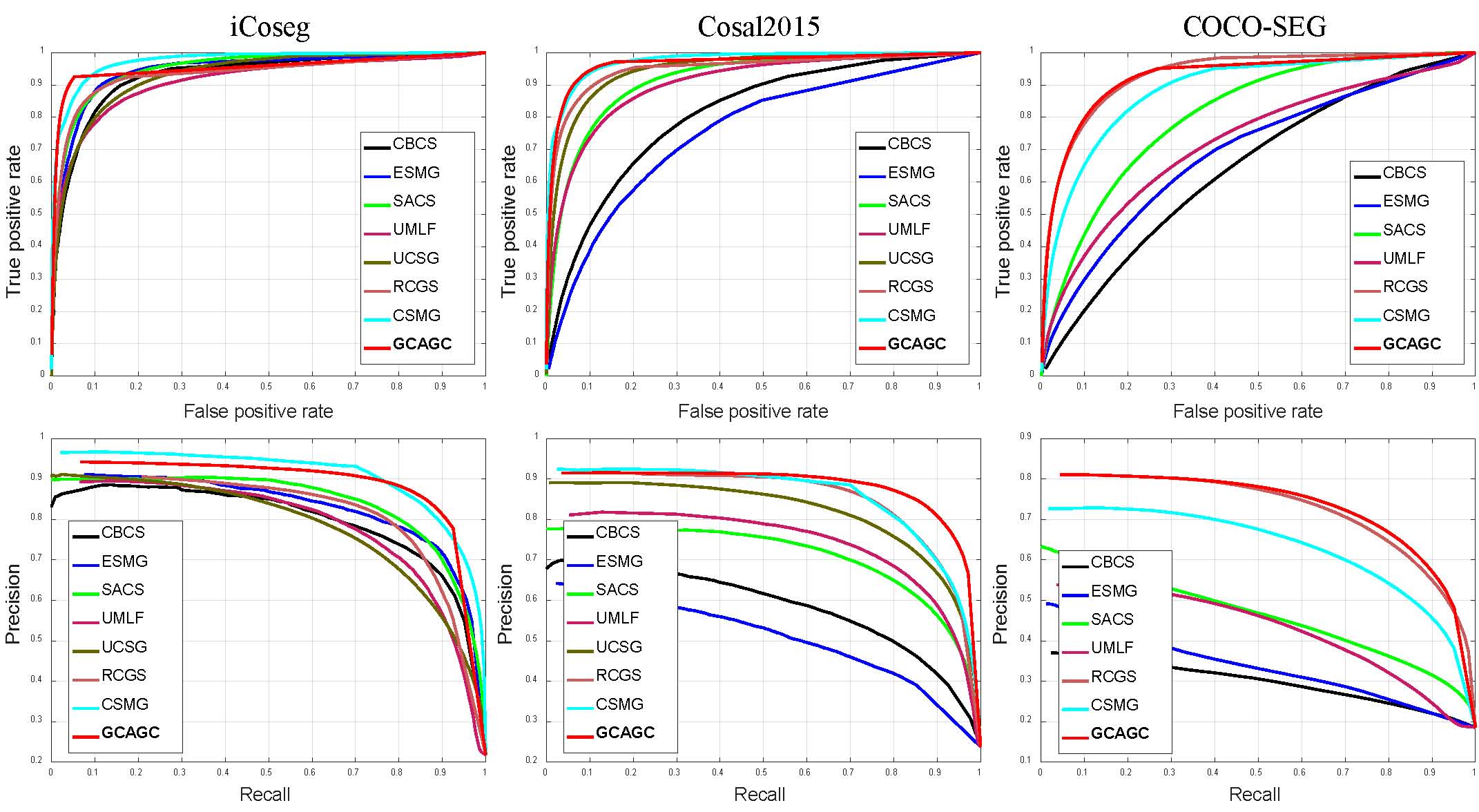}
\end{tabular}
\end{center}
   \caption{Comparisons with state-of-the-art methods in terms of PR and ROC curves on three benchmark datasets}
\label{fig:PRcurves}
\end{figure*}
\begin{table*}[t]
\small
\caption{Statistic comparisons of our GCAGC with the other state-of-the-arts. \textcolor[rgb]{1.00,0.00,0.00}{\textbf{Red}} and \textcolor[rgb]{0.00,0.00,1.00}{{\textbf{blue}}} bold fonts indicate the best and second best performance, respectively. }
\setlength{\tabcolsep}{1.4mm}
\begin{center}
\begin{tabular}{|l||c c c c|c c c c|c c c c|}
\hline
\multirow{2}{*}{Methods}& \multicolumn{4}{c|}{iCoseg} & \multicolumn{4}{c|}{Cosal2015}& \multicolumn{4}{c|}{COCO-SEG} \\
\cline{2-13}
          &AP$\uparrow$ & ${F_\beta}$$\uparrow$ & $S_{m}$$\uparrow$ & MAE$\downarrow$ & AP$\uparrow$ & ${F_\beta}$$\uparrow$ & $S_{m}$$\uparrow$ & MAE$\downarrow$ & AP$\uparrow$ & ${F_\beta}$$\uparrow$ & $S_{m}$$\uparrow$ & MAE$\downarrow$ \\
\hline
CBCS~\cite{fu2013cluster}    &0.7965 &0.7408 &0.6580 &0.1659 &0.5859 &0.5579 &0.5439 &0.2329  & 0.3043  & 0.3050 & 0.4710 & 0.2585\\
CSHS~\cite{liu2014co}        &0.8454 &0.7549 &0.7502 & 0.1774 &0.6198 &0.6210 &0.5909 &0.3108 & - & - & - & - \\
ESMG~\cite{li2015efficient}    &0.8336 &0.7773 &0.7677 &0.1261 &0.5116 &0.5120 &0.5446 &0.2581  & 0.3387  &0.3592  & 0.4931 & 0.2349\\
SACS~\cite{cao2014self}      &0.8399 &0.7978 &0.7523 &0.1516 &0.7076 &0.6927 &0.6938 &0.1920  &0.4176 & 0.4234 & 0.5229 & 0.3271   \\
CODW~\cite{zhang2015co}      &0.8766 &0.7990 &0.7500 & 0.1782 &0.7437 &0.7051 &0.6473 &0.2733 & - & - & - & -\\
DIM~\cite{zhang2015cosaliency} & 0.8773 & 0.7919 & 0.7583 & 0.1739 & 0.6305 & 0.6162 & 0.5907 & 0.3123 & 0.3043 & 0.3353 & 0.4572 & 0.3871 \\
UMLF~\cite{han2017unified}        & 0.7881 & 0.7148 & 0.7033 & 0.2389 & 0.7444 & 0.7016 & 0.6604 & 0.2687 & 0.4347 & 0.4309 & 0.4872 & 0.3953  \\
UCSG~\cite{hsu2018unsupervised}    &\color{red}\textbf{0.9112} &0.8503 &0.8200  & 0.1182 &0.8149 &0.7589 &0.7506 &0.1581 & - & - &- &-\\
RCGS~\cite{wang2019robust}         &0.8269  &0.7730 &0.7810 & \color{blue}\textbf{0.0976} &\color{blue}\textbf{0.8573} &0.8097 &\color{blue}\textbf{0.7959} &\color{blue}\textbf{0.0999} &\color{blue}\textbf{0.7309}  &\color{blue}\textbf{0.6814}  &\color{blue}\textbf{0.7185} &\color{blue}\textbf{0.1239} \\
CSMG~\cite{zhang2019co}            &\color{blue}\textbf{0.9097}  &\color{blue}\textbf{0.8517} &\color{red}\textbf{0.8208} & 0.1050 &0.8569 &\color{blue}\textbf{0.8216} &0.7738 &0.1292 & 0.6309 & 0.6208 &0.6517 &0.1461\\
\textbf{GCAGC}                     & 0.8867 & \color{red}\textbf{0.8532} &\color{blue}\textbf{0.8205}&  \color{red}\textbf{0.0757}& \color{red}\textbf{0.8799} & \color{red}\textbf{0.8428} & \color{red}\textbf{0.8224} & \color{red}\textbf{0.0890}  & \color{red}\textbf{0.7323} & \color{red}\textbf{0.7092} & \color{red}\textbf{0.7294} & \color{red}\textbf{0.1097}\\
\hline
\end{tabular}
\end{center}
\label{table}
\end{table*}
\subsection{Datasets and Evaluation Metrics}
\label{sec:setup}
We conduct extensive evaluations on three popular datasets including iCoseg~\cite{batra2010icoseg}, Cosal2015~\cite{zhang2015co} and COCO-SEG~\cite{wang2019robust}.
Among them, iCoseg is the most widely used dataset with totally $38$ groups of $643$ images, among which the common objects in one group share similar appearance or semantical characteristics, but have various pose or color changes.
Cosal2015 is a large-scale dataset which is consist of $2,015$ images of $50$ categories, and each group suffers from various challenging factors such as complex environments, occlusion issues, target appearance variations and background clutters, \textit{etc}. All these increase the difficulty for accurate co-saliency detection.
Recently, to meet the urgent requirement of large-scale training set for deep-learning-based co-saliency detection approaches, COCO-SEG has been proposed which are selected from the COCO2017 dataset~\cite{lin2014microsoft}, of which $200,000$ and $8,000$ images are for training and testing respectively from all $78$ categories.
%

We compare our GCAGC method with existing state-of-the-art algorithms in terms of $6$ metrics including the precision-recall (PR) curve~\cite{zhang2018review}, the receive operator characteristic (ROC) curve~\cite{zhang2018review}, the average precision (AP) score~\cite{zhang2018review}, F-measure score $F_{\beta}$~\cite{zhang2018review}, S-measure score $S_m$~\cite{fan2017structure} and Mean Absolute Error (MAE)~\cite{wang2019robust}.
%
%
%
\subsection{Comparisons with State-of-the-arts}\label{sec:comparisions}
We compare our GCAGC approach with $10$ state-of-the-art co-saliency detection methods including  CBCS~\cite{fu2013cluster}, CSHS~\cite{liu2014co},  ESMG~\cite{li2015efficient}, SACS~\cite{cao2014self}, CODW~\cite{zhang2015co},
DIM~\cite{zhang2015cosaliency}, UMLF~\cite{han2017unified}, UCSG~\cite{hsu2018unsupervised}, RCGS~\cite{wang2019robust}, CSMG~\cite{zhang2019co}.
For fair comparisons, we directly report available results released by authors or reproduce experimental results by the public source code for each compared method.
%

\begin{table}[t]
\footnotesize
\caption{Ablative studies of our model on iCoseg and Cosal2015. Here GCAGC-N, GCAGC-M, GCAGC-P denote our GCAGC in absence of AGCN, AGCM and the projection matrices $\textbf{P}$ in (\ref{eq:Ak}), respectively. \textcolor[rgb]{1.00,0.00,0.00}{\textbf{Red}} bold font indicates the best performance.}
\setlength{\tabcolsep}{1.0mm}
\begin{center}
\begin{tabular}{|c||c|c|c|c|c|}
\hline
Datasets&   &GCAGC-N  & GCAGC-M  &GCAGC-P &GCAGC   \\
\hline
        &AP$\uparrow$         &0.8799&0.8606&0.8796&\color{red}\textbf{0.8867}\\
iCoseg  &${F_\beta}$$\uparrow$&0.8504&0.8123&0.8463&\color{red}\textbf{0.8532}\\
        &$S_{m}$$\uparrow$    &0.8175&0.8203&0.8122&\color{red}\textbf{0.8205}\\
        & MAE$\downarrow$       &0.0831&0.0796&0.0790&\color{red}\textbf{0.0757}\\
\hline
           &AP$\uparrow$         &0.8577&0.8779&0.8737&\color{red}\textbf{0.8799}\\
Cosal2015  &${F_\beta}$$\uparrow$ &0.8156&0.8373&0.8375&\color{red}\textbf{0.8428}\\
           &$S_{m}$$\uparrow$    &{0.8167}& 0.8145 &0.8156& \color{red}\textbf{0.8224}\\
           & MAE$\downarrow$       &0.0967&0.0901&0.0851&\color{red}\textbf{0.0890}\\
\hline
\end{tabular}
\end{center}
\label{fig:ablation}
\end{table}
%
\textbf{Qualitative Results.}
Figure~\ref{fig:Comparison1} shows some visual comparison results with 4 state-of-the-art methods including CBCS~\cite{fu2013cluster}, ESMG~\cite{li2015efficient}, CSMG~\cite{zhang2019co} and RCGS~\cite{wang2019robust}.
Our GCAGC can achieve better co-saliency results than the other methods when the co-salient targets suffer from significant appearance variations, strong semantic interference and complex background clutters.
In Figure~\ref{fig:Comparison1}, the two left groups of images are selected from iCoseg.
Among them, for the group of \textit{Red Sox Players}, the audience in the background share the same semantics with those foreground co-salient players,
which makes it very difficult to accurately differentiate them.
Notwithstanding, our GCAGC can accurately highlight the co-salient players due to its two-steps filtering processing from GC filtering to graph clustering that can well preserve spatial consistency while effectively reducing noisy backgrounds.
However, the other compared methods cannot achieve satisfying results, which contain either some noisy backgrounds (see the middle columns of RCGS, ESMG, CBCS) or the whole intra-salient areas including non-co-salient regions (see the left-most column of RCGS, the left-fourth columns of ESMG and CBCS).
The co-saliency maps in the middle groups (\textit{Apple} and \textit{Monkey}) are generated from the image groups selected from Cosal2015.
The \textit{Apple} group suffers from the interferences of other foreground semantic objects such as hand and lemon while the \textit{Monkey} group undergoes complex background clutters.
It is obvious that our GCAGC can generate better spatially coherent co-saliency maps than the other methods (see the two bottom rows of ESMG and CBCS, the left-most columns of RCGS and CSMG).
%
%
The two right-most groups are selected from COCO-SEG, which contain a variety of challenging images with targets suffering from the interferences of various different categories and complicate background clutters.
Notwithstanding, our GCAGC can accurately discover the co-salient targets even when they suffer from extremely complicate background clutters (see the \textit{Broccoli} group).
The experimental results show that our GCAGC can achieve favorable performance against various challenging factors, validating the effectiveness of our GCAGC model that can adapt well to a variety of complicate scenarios.

\textbf{Quantitative Results.} Figure~\ref{fig:PRcurves} shows the PR and the ROC curves of all compared methods on three benchmark datasets.
We can observe that our GCAGC outperforms the other state-of-the-art methods on three datasets. Especially, all the curves on the largest and most challenging Cosal2015 and COCO-SEG are much higher than the other methods.
Meanwhile, Table~\ref{table} lists the statistic analysis, among which the RCGS is a representative end-to-end deep-learning-based method that achieves state-of-the-art performance on both Cosal2015 and COCO-SEG with the F-scores of $0.8097$ and $0.6814$, respectively.
Our GCAGC achieves the best F-scores of $0.8428$ and $0.7092$ on Cosal2015 and COCO-SEG, respectively, outperforming the second best-performing CSMG by $3.31\%$ on Cosal2015 and RCGS by $2.78\%$ on COCO-SEG.
All the qualitative results further demonstrate the effectiveness of jointly learning the GCGAC model that is essential to accurate co-saliency detection.
%
%
%
\subsection{Ablative Studies}
\label{sec:ablation}
%
%
Here, we conduct ablative studies to validate the effectiveness of the proposed two modules (AGCN and AGCM) and the adaptive graph learning strategy in the AGCN.
%
Table~\ref{fig:ablation} lists the corresponding quantitative statistic results in terms of AP, ${F_\beta}$, $S_{m}$ and MAE.

First, without AGCN, the GCAGC-N shows obvious performance drop on Cosal2015 in terms of all metrics, especially for both AP and $F_\beta$, where the former drops from $0.8799$ to $0.8577$ by $2.22\%$ and the latter drops from $0.8428$ to $0.8156$ by $2.72\%$.
Besides, the performance of GCAGC-N on iCoseg also suffers from drop in terms of all metrics.

Second, without AGCM, the GCAGC-M suffers from obvious performance drop in terms of all metrics on both datasets, especially for AP and $F_{\beta}$ on iCoseg, where the AP score and the $F_\beta$ decline from $0.8867$ to $0.8606$ by $2.61\%$ and from $0.8532$ to $0.8123$ by $4.09\%$, respectively.
The results validate the effectiveness of the proposed AGCM that can well discriminate the co-objects from all the salient foreground objects to further boost the performance.

Finally, without adaptive graph learning in AGCN, all metrics in GCAGC-P have obvious decline on both datasets, further showing the superiority of proposed AGCN to learn an adaptive graph structure tailored to the co-saliency detection task compared with the fixed graph design in the vanilla GCNs~\cite{kipf2016semi}.
\section{Conclusion}
This paper has presented an adaptive graph convolutional network with attention graph clustering for co-saliency detection, mainly including two key designs: an AGCN and an AGCM.
The AGCN has been developed to extract long-range dependency cues to characterize the intra- and inter-image correspondence. Meanwhile, to further refine the results of the AGCN, the AGCM has been designed to discriminate the co-objects from all the salient foreground objects in an unsupervised fashion.
Finally, a unified framework with encoder-decoder structure has been implemented to jointly optimize the AGCN and the AGCM in an end-to-end manner.
Extensive evaluations on three largest and most challenging benchmark datasets including iCoseg, Cosal2015 and COCO-SEG have demonstrated superior performance of the proposed method over the state-of-the-art methods in terms of most metrics.
{\small
\bibliographystyle{ieee}
\bibliography{egbib}
}


\end{document}